\def\year{2021}\relax
\documentclass[letterpaper]{article} 
\usepackage{aaai21}  
\usepackage{times}  
\usepackage{helvet} 
\usepackage{courier}  
\usepackage[hyphens]{url}  
\usepackage{graphicx} 
\usepackage{natbib}  
\usepackage{caption} 
\usepackage{enumitem}
\usepackage{comment}

\usepackage{graphicx}
\usepackage{subcaption}

\usepackage{url}
\usepackage{booktabs}
\usepackage{todonotes}

\newcommand{\participation}{\textsc{Participation}}
\newcommand{\acceptance}{\textsc{Acceptance}}
\newcommand{\propagation}{\textsc{Propagation}}
\newcommand{\posprovocation}{\textsc{PosProvocation}}
\newcommand{\negprovocation}{\textsc{NegProvocation}}

\newcommand{\newcite}[1]{\cite{#1}}

\title{How Metaphors Impact Political Discourse: \\A Large-Scale Topic-Agnostic Study Using Neural Metaphor Detection}
\author{Vinodkumar Prabhakaran\textsuperscript{\rm 1}, 
        Marek Rei\textsuperscript{\rm 2},
        Ekaterina Shutova\textsuperscript{\rm 3} \\
}
\affiliations {
    \textsuperscript{\rm 1}Google Research, USA \\
    \textsuperscript{\rm 2}Imperial College London, UK \\
    \textsuperscript{\rm 3}University of Amsterdam, Netherlands \\
     vinodkpg@google.com, marek.rei@imperial.ac.uk, e.shutova@uva.nl
}

\begin{document}

\maketitle
\begin{abstract}
\begin{quote}
Metaphors are widely used in political rhetoric as an effective framing device. While the efficacy of specific metaphors such as the war metaphor in political discourse has been documented before, those studies often rely on small number of hand-coded instances of metaphor use. Larger-scale topic-agnostic studies are required to establish the general persuasiveness of metaphors as a device, and to shed light on the broader patterns that guide their persuasiveness. In this paper, we present a large-scale data-driven study of metaphors used in political discourse. We conduct this study on a publicly available dataset of over 85K posts made by 412 US politicians in their Facebook public pages, up until Feb 2017. Our contributions are threefold: we show evidence that metaphor use correlates with ideological leanings in complex ways that depend on concurrent political events such as winning or losing elections; we show that posts with metaphors elicit more engagement from their audience overall even after controlling for various socio-political factors such as gender and political party affiliation; and finally, we demonstrate that metaphoricity is indeed the reason for increased engagement of posts, through a fine-grained linguistic analysis of metaphorical vs. literal usages of 513 words across 70K posts. 
\end{quote}
\end{abstract}


\noindent Metaphorical expressions arise in the presence of systematic metaphorical associations, or \textit{conceptual metaphors}, mapping one concept or domain to another \cite{LakoffAndJohnson}. For instance, when we talk about ``\textit{curing} juvenile delinquency'' or ``\textit{diagnosing} corruption'', we view \textit{crime} (the target domain) in terms of a \textit{disease} (the source domain) and map various elements of the \textit{disease} knowledge system to our reasoning about crime. Since recognizing metaphorical expressions is critical in order to correctly interpret their intended meanings, computational approaches to metaphor detection has long been an active area of research in Natural Language Processing (NLP) \cite{shutova2010models}. While recent research has achieved great advances in  automatic metaphor detection performance \cite{DarioACL2016,Bulat2017,MarekEMNLP2017,gao2018neural,Dankers2019}, not much research has investigated whether metaphor detection, when employed at scale, could help answer some of the fundamental questions in social sciences about metaphors and their ability to shape the public discourse.

Social scientists have demonstrated the importance of metaphors in political rhetoric --- their role as a \textit{framing} technique \cite{Lakoff1991,Tannen,Entman2003}, their tendency to elicit stronger emotions than their literal counterparts \cite{citron2014metaphorical},  and their effectiveness in influencing decision-making \cite{ThibodeauBoroditsky}. For instance, metaphors help with the framing of an issue by selecting and emphasising its facets that reinforce a particular point of view \cite{Lakoff1991,Musolff,LakoffWehling}. Discussing \textit{war} as a \textit{competitive game} emphasizes the victory vs. defeat aspect of war, while neglecting its human cost, a strategy politicians could use to arouse a pro-war sentiment in the public \cite{Lakoff1991}. While the role metaphors play in shaping the public discourse has been documented before, these qualitative studies are often limited to a few specific domains or topics (e.g., the \textit{war} domain above) and the metaphors used in them. 

In this paper, aided by a neural network based metaphor detection approach, we present a large-scale domain-agnostic study of the effects of metaphors in political discourse. Our study is conducted on a dataset of 85K Facebook status message posts made by 412 US politicians over a period of around nine years. Our three main findings are:
\begin{itemize}
\item We find that politicians' rate of metaphor use is correlated with their ideological leanings in complex ways that depend on contemporaneous political events. Specifically, our analysis found that Democratic politicians used significantly more metaphors during the three months immediately after the 2016 election loss, compared to themselves in prior months, as well as to Republican politicians during the same period of time or prior months.  
\item We find that politicians' posts that include metaphorical language tend to have significantly higher engagement from their audience. The scale of our study enabled us to control for various socio-political factors, such as gender and party affiliation, strengthening our findings compared to prior studies.
\item We also conduct a finer-grained linguistic analysis that reveals (1) that metaphorical uses of source domain words lead to a higher engagement than their literal usage, and (2) aspects of target domains (i.e. specific political issues) that receive a greater engagement when described metaphorically, as compared to literally. 
\end{itemize}
To our knowledge, this is the first computational study of this scale on the impact of metaphor use in political communication in a topic-agnostic manner.

\section{Related Work}

Computational social scientists have demonstrated the impact language use has on the reaction it can evoke on its audience in large-scale data-driven studies. For instance, \cite{danescu2012you} demonstrated that lexical distinctiveness --- i.e., using less common word choices, while relying on common syntactic patterns --- boosts the memorability of movie quotes. 
Others have studied how linguistic features and techniques shape the perception of helpfulness of online product reviews \cite{danescu2009opinions}, persuasiveness of funding requests \cite{yang2019let}, and virality of social media posts \cite{guerini2012exploring,tan2014effect}. Studies have also looked into the effects of usage of titles, names etc. \cite{lakkaraju2013s} for better social media targeting.
Computational linguistics analysis has also been employed to study various aspects of political speech, for instance, persuasiveness \cite{guerini2008trusting}, charisma \cite{rosenberg2009charisma}, and power dynamics \cite{prabhakaran-etal-2013-upper,prabhakaran-etal-2014-staying}.
Our work is situated within this line of computational inquiries that look into the social effects of linguistic patterns used in political speech. Specifically, we study the use of metaphors in political discourse.



Political science offers a rich literature on the value of metaphor in political rhetoric. Studies have
documented the effectiveness of metaphors as a \textit{framing} technique \cite{Lakoff1991,Tannen,Entman2003}
that shapes public sentiment by exposing desired aspects of an issue, while seamlessly concealing the less desired ones \cite{Musolff,LakoffWehling} (as in the \textit{war} example above). Metaphors have also been found to express and elicit stronger emotions than literal expressions \cite{Crawford2009,citron2014metaphorical,mohammad2016metaphor}; e.g., the phrase ``a \textit{tsunami} of immigrants'' portrays immigration as a threat and spurs fear, as compared to ``\textit{high numbers} of immigrants''. 

Studies have also documented differences in the metaphors used across communities with different political views. For instance, \newcite{LakoffMoralPolitics} shows that two conflicting instantiations of the \textsc{nation is a family} metaphor, using \textit{nurturing-parent} vs. \textit{strict-father} family models, explain the liberal-conservative divide in the US politics.
\newcite{LakoffWehling} show that the two models are consistent with both the parties' rhetoric and their policies, and can be used to promote liberal vs. conservative values.  
Taking a step further, \newcite{ThibodeauBoroditsky} investigated how metaphors affect our decision-making. In their experiment, two groups of subjects were primed by two different metaphors for \textit{crime}: \textit{crime is a virus} vs. \textit{crime is a beast} and then asked how crime should be tackled. They found that the first group tended to opt for preventive measures and the second group for punishment-oriented ones. According to the authors, their results demonstrate the influence that metaphors have on how we conceptualize and act with respect to societal issues. 

Drawing inspiration from this line of research, we demonstrate through a large-scale domain-agnostic study that automatic metaphor detection could help us better understand the broader patterns that guide the impact of metaphor in political discourse. 

\section{Data}
\label{sec_data}

We use the dataset of Facebook posts made by US politicians in their public Facebook pages created by \cite{voigt-etal-2018-rtgender}. The dataset contains over 399K posts, all made in English, by 402 politicians who were either a US House Representative or a US Senator at the time of data collection. The dataset captures the politicians' binary-gender (306 male and 96 female) and party affiliation. The dataset also contains the statistics on the number of reactions each post got (\textit{Like}, \textit{Love}, \textit{Haha}, \textit{Wow}, \textit{Angry}, and \textit{Sad}), the number of times it was shared, and the comments it received. 

Of all the posts, about 85K were text-only status messages. We limit our study to this subset of the data since we are interested in understanding the impact of language use specifically, and the presence of images, videos and links makes it difficult to tease that apart. 
%
%
%
%
As our dataset contains only US politicians, our study is limited to the English language and to the sociopolitical context of the US. Follow up work may investigate whether our findings extend to other geographic regions, cultural contexts, and languages. 



\section{Metaphor Identification}
\label{sec_classifier}

Metaphors are frequently used in political discourse. For instance, the following three sentence fragments from our dataset demonstrate literal vs. metaphorical description of the same concept, \textit{economy}:

\begin{quote}
    \textit{My top priority is to improve the local economy and create jobs.} (1)
\end{quote}

\begin{quote}
    \textit{it's time to jumpstart the economy and put it to work for the middle-class} (2)
\end{quote}

\begin{quote}
    \textit{we need to stop stifling our own economy} (3)
\end{quote}

\noindent In (1), the politician is describing the term economy literally by using the verb \textit{improve}, whereas in (2) and (3), the politician uses the verbs \textit{jumpstart} and \textit{stifle} to metaphorically describe the term economy. The verb \textit{jumpstart} evokes the frame of economy being like a car that has broken down, while the verb \textit{stifle} evokes the frame of economy being a person who is suffocated or restrained. 
We are interested in studying the patterns associated with metaphor use in political discourse, and whether they result in increased engagement with the audience. We start by automatically identifying metaphors in our data.

\subsection{Background}


Corpus linguistics studies have long documented that a large majority of metaphors involve metaphorically-used verbs, with adjectives being another frequent category \cite{cameron2003metaphor,ShutovaThesis}. Motivated by their frequency in metaphorical constructions, we focus on metaphorically-used verbs and adjectives in our study. Our decision is also in part driven by the availability of resources for these types of metaphors \cite{Boytsov2014,DarioACL2016,mohammad2016metaphor}. Specifically, we study verb-noun metaphors in both subject-verb and verb-direct object phrases. In addition, we also study adjective-noun metaphors, which are relatively rare in our corpus compared to metaphorical uses of verbs (similar to previous corpus studies mentioned above).

Early research on automatic metaphor detection in text used supervised learning methods with hand-engineered features, encoding lexico-syntactic patterns \citep{mohler-EtAl:2013:Meta4NLP,DunnCicling2013,Boytsov2014,BeigmanKlebanovACL2016}; topical structure \citep{heintz-EtAl:2013:Meta4NLP,strzalkowski-EtAl:2013:Meta4NLP}; topic-based frame templates \cite{jang2017finding,jang2017computational}; selectional preferences \citep{wilks-EtAl:2013:Meta4NLP,shutova-2013-metaphor}; and psycholinguistic features e.g. abstractness \citep{Turney2011} and imageability \citep{strzalkowski-EtAl:2013:Meta4NLP}. More recent approaches use distributed representations and neural models \cite{DarioACL2016,Bulat2017,MarekEMNLP2017,gao2018neural,Dankers2019}. 
%

In this work, we employed the neural framework of \newcite{MarekEMNLP2017}, where a metaphor classifier takes a word pair, such as (\textit{cure}, \textit{crime}) or (\textit{blind}, \textit{hope}), as input and returns a score between $0$ and $1$ indicating the likelihood that the pair is a metaphorical rather than literal usage.
The method builds on the idea that cosine similarity between the embeddings of two words is indicative of metaphoricity \cite{ShutovaNAACL2016}, and extends it to a supervised version of similarity optimized for metaphor detection using annotated examples. 


We opted for the model of \newcite{MarekEMNLP2017} for a number of reasons. This model achieves (near) state-of-the-art performance on two metaphor identification benchmarks: \cite{Boytsov2014} and \cite{mohammad2016metaphor}. Since it is applied to individual word pairs, it is likely to be less sensitive to domain shifts than the models that process complete sentences. 
This domain robustness is critical for our study, since in-domain (i.e., in political discourse) metaphor annotations are not available.
In contrast, alternative models of metaphor \cite{gao2018neural,Dankers2019} were trained on the VU Amsterdam Metaphor Corpus \cite{SteenMIPVU}, which focuses on historical aspects of metaphor and includes a high number of highly conventionalised and dead metaphors (e.g. prepositions used metaphorically). Such metaphors are less interesting in our study as we focus on contemporary and productive metaphor use, which are best captured by the model of \cite{MarekEMNLP2017} due to their choice of training data. Furthermore, the code and trained models of \cite{MarekEMNLP2017} are publicly available, which facilitates replicability of our experiments.

\subsection{Model and Training}

We now
describe our metaphor detection model, along with the experiments we conducted, as well as the configurations we used for training. 
%
We train two metaphor classifiers following the architecture described by \newcite{MarekEMNLP2017}.
The model is designed to take a word pair, such as (\textit{cure}, \textit{crime}) or (\textit{blind}, \textit{hope}), as input and return a score indicating whether the corresponding phrase is metaphorical or literal. 

The network takes as input two words and maps them to their corresponding word embeddings $x_1$ and $x_2$. 
Next, the representation of the first word is used to apply gating on the second word:
\begin{equation}
     g = \sigma(W_g x_1 + b_g)
\end{equation}
\begin{equation}
    \widetilde{x}_2 = g \odot x_2
\end{equation}

\noindent where $W_g$ is a weight matrix, $b_g$ is a bias vector and $\odot$ is element-wise multiplication.
This is designed to mimic the interaction between the domains of the two words and how the meaning of the target word can change in a metaphorical phrase depending on the context in which it appears.

Each word representation is then mapped to a new space using a $tanh$ layer:
\begin{equation}
    z_1 = tanh(W_{z_1} x_1 + b_{z_1})
\end{equation}
\begin{equation}
    z_2 = tanh(W_{z_2} \widetilde{x}_2 + b_{z_2})
\end{equation}

\noindent where $W_{z_1}$ and $W_{z_2}$ are weight matrices, $b_{z_1}$ and $b_{z_2}$ are bias vectors.

The original input embeddings are trained with the skip-gram objective \cite{Mikolov2013a} on Wikipedia. 
By learning this explicit mapping, we allow the model to find a new space that takes advantage of the pre-training but also better captures the differences between metaphorical and literal meanings. Importantly, the mappings are position-specific; each word in the pair is treated differently depending on its linguistic role in the phrase.

Finally, the two word representations are combined through element-wise multiplication and passed through a hidden layer, which allows the model to learn a supervised approximation of cosine similarity for this task. The output prediction is made through a logistic activation function.

\begin{equation}
    d = tanh(W_d (z_1 \odot z_2) + b_d)
\end{equation}

\begin{equation}
    \widetilde{y} = \sigma(W_y d + b_y)
\end{equation}

\noindent where $W_d$ and $W_y$ are trainable weight matrices, $b_d$ and $b_y$ are bias vectors and a high output value $\widetilde{y}$ indicates a metaphorical phrase.

The model parameters are optimized using a hinge loss described by \newcite{MarekEMNLP2017}, which only trains using datapoints selected based on their distance from the decision boundary.
We train two versions of this architecture:
\begin{enumerate}
    \item Detecting metaphors in adjective--noun phrases, trained using the combined datasets of \newcite{Boytsov2014} and \newcite{DarioACL2016}.
    \item Detecting metaphors in verb--subject and verb--direct object word pairs, trained using the dataset from \newcite{mohammad2016metaphor}.
\end{enumerate}

These two models are applied at the same time in order to detect metaphors of both types in Facebook posts. While the architecture by \newcite{MarekEMNLP2017} also used attribute-based vectors as input, we train the model only with skip-gram embeddings. The attribute vectors are very large and need to be pre-calculated for a specific vocabulary. By using skip-gram embeddings instead, we are able to apply the model to a wider range of text.

The 100-dimensional input word embeddings were pre-trained on Wikipedia and are kept fixed while the rest of the model is optimized, to prevent overfitting.
The mapped representations $z_1$ and $z_2$ have size $300$, layer $d$ has size $50$.
The input word embeddings have a vocabulary size of $184,805$, resulting in a total of $18,480,500$ fixed parameters. 
The network itself contains $85,801$ trainable parameters.
The network was optimized using AdaDelta \cite{Zeiler2012} on a single Titan Xp GPU and training was stopped based on performance on the respective development splits for each dataset.

\subsection{Evaluation}

Metaphor detection on the \newcite{mohammad2016metaphor} dataset is normally evaluated with cross-validation, but in order to train a single model that can be applied on the Facebook dataset we split this dataset into 517/65/65 instances for train/dev/test. To evaluate the classifier for adjective-noun metaphors the combined training data contains 9,880 examples and the \newcite{Boytsov2014} dev/test sets contain 200 examples each.
The verb--argument classifier has $73.85\%$ accuracy on the dev set and  $75.38\%$ accuracy on the test set of the \newcite{mohammad2016metaphor} dataset. The adjective--noun classifier achieves $85.71\%$ F-score on the \newcite{Boytsov2014} dev set and $86.73\%$ F-score on the test set.


\subsection{Metaphoricity of Facebook Posts}

We parse the text of each Facebook post in our data
using the spaCy tool \cite{honnibal-johnson:2015:EMNLP} 
to identify the candidate verb--subject, verb--object, and adjective--noun pairs, which are then passed to the corresponding classifier. 
We use a threshold of 0.7 for the score (ranging between 0 and 1) for a word pair to be considered a metaphor (as used in the evaluation described above).\footnote{Experiments using lower thresholds (0.5 and 0.6) were conducted, which also obtained similar results.}
We define the \textit{metaphoricity} of a post as the number of metaphors present in it. 
Only 17\% of posts in our corpus had a non-zero \textit{metaphoricity}, and 96\% of them were between 1 and 3. Note that longer posts are likely to contain more metaphors; so we control for post length in all our analyses.

\section{Study 1: Usage Analysis}
\label{sec_usage}

\begin{figure}[t]
\centering
\begin{subfigure}{.45\linewidth}
  \centering
  \includegraphics[width=.99\linewidth]{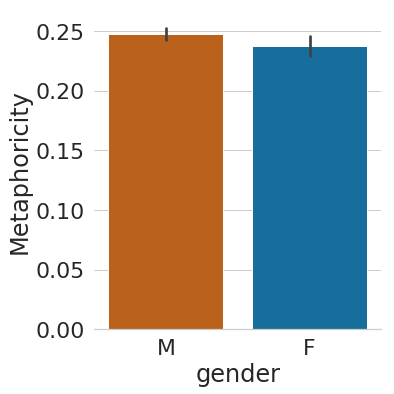}
  \caption{}
  \label{fig_gp:sfig1}
\end{subfigure}%
\begin{subfigure}{.45\linewidth}
  \centering
  \includegraphics[width=.99\linewidth]{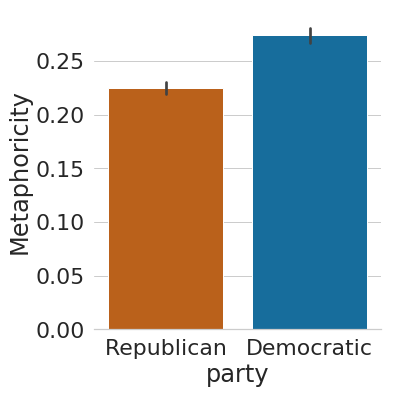}
  \caption{}
  \label{fig_gp:sfig2}
\end{subfigure}
\caption{Box plots showing differences in metaphor use across gender and party affiliation}
\label{fig:fig_genderparty}
\end{figure}

Does the frequency of metaphors used by politicians depend on their gender and political party affiliation? Figure~\ref{fig:fig_genderparty} shows the box plots of differences in metaphor use across gender (binary) and party affiliation.\footnote{We count Bernie Sanders (Independent) among Democrats, based on his voting records.} We found no significant difference across gender, while we found that Democratic politicians used significantly more metaphors in our data.

Our next question was how the political events may shape the use of metaphors. For instance, winning or losing an election may influence one's need to evoke emotional responses. To answer this question, we analyzed the metaphor use differences in the months leading up to and after the Nov 2016 US elections. For this subset of the study, we limit our analysis to only 8K posts made between 07 Feb 2016 and 06 Feb 2017, divided into 4 quarters --- three quarters ($Q_{-3}$, $Q_{-2}$, and $Q_{-1}$) prior to the election, and one quarter ($Q_{+1}$) after the election. 

\begin{figure}
    \centering
    \includegraphics[width=.8\linewidth]{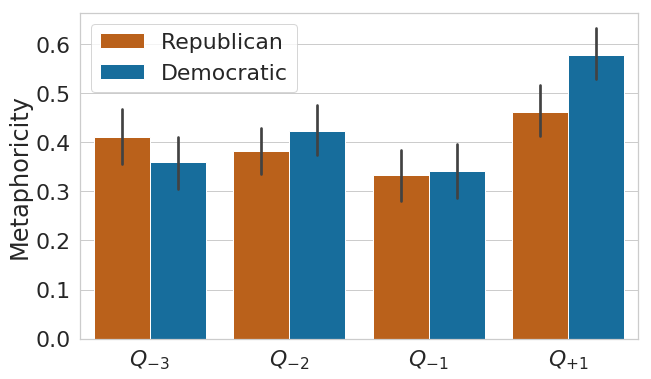}
    \caption{Use of metaphors along party lines in different quarters relative to the 2016 US Elections}
    \label{fig:temporal}
\end{figure}

After controlling for post length, we found no significant difference in metaphoricity
across gender (b=.025, t(1.411), p=0.158) or party (b=.0260, t(1.648), p=0.099), reconfirming the results from above. 
However, the difference in metaphor use along party affiliation was present on in $Q_{+1}$.
A one-way ANOVA showed significant differences ($p<0.001$) in metaphoricity of posts across quarters, while a two-way ANOVA revealed a significant interaction ($p<0.05$) between the party and the quarter, suggesting that Democrats and Republicans differed in their metaphor use across the quarters. 
Figure~\ref{fig:temporal} shows average metaphoricity of posts across both parties in different quarters. The main significant shift was in terms of Democratic politicians using more metaphors in $Q_{+1}$ than other quarters and than Republicans. In fact, a post-hoc analysis using Tukey's pairwise HSD test revealed that it was the only group of posts that significantly differed from all other groups. 

We speculate that this difference in behavior between Republicans and Democrats is caused by the increased emotions on the Democratic side in response to the election loss. While establishing the underlying reason for this difference is outside the scope of this paper, this finding demonstrates that it is important to account for the contemporaneous political wins and setbacks while analysing differences in language use across party lines. 

\begin{figure*}[t]
\centering
\begin{subfigure}{.28\linewidth}
  \centering
  \includegraphics[width=.99\linewidth]{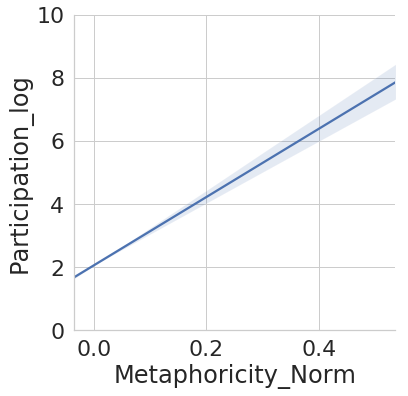}
  \caption{}
  \label{fig:sfig1}
\end{subfigure}%
\begin{subfigure}{.28\linewidth}
  \centering
  \includegraphics[width=.99\linewidth]{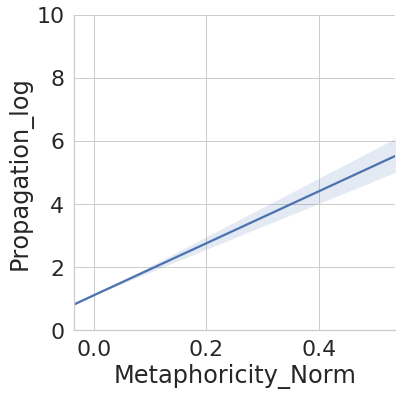}
  \caption{}
  \label{fig:sfig2}
\end{subfigure}
\begin{subfigure}{.28\linewidth}
  \centering
  \includegraphics[width=.99\linewidth]{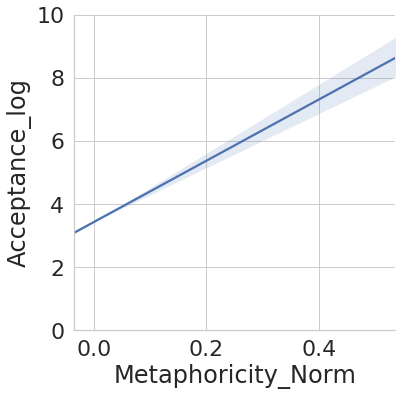}
  \caption{}
  \label{fig:sfig3}
\end{subfigure}\\
\begin{subfigure}{.28\linewidth}
  \centering
  \includegraphics[width=.99\linewidth]{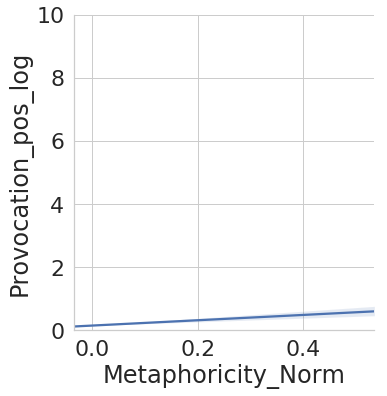}
  \caption{}
  \label{fig:sfig4}
\end{subfigure}
\begin{subfigure}{.28\linewidth}
  \centering
  \includegraphics[width=.99\linewidth]{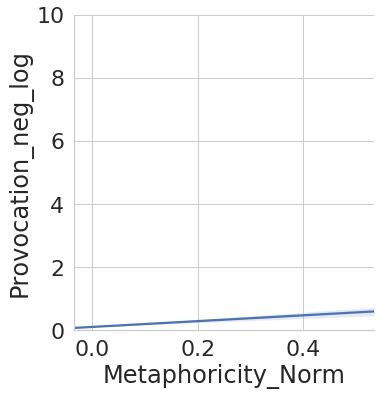}
  \caption{}
  \label{fig:sfig5}
\end{subfigure}
\caption{Regression plots showing post-level engagement differences depending on Metaphoricity. The metaphoricity of a post is normalized w.r.t. post length (number of words), denoted by Metaphoricity\_Norm}
\label{fig:fig_lmplots}
\end{figure*}

\section{Study 2: Engagement Analysis}
\label{sec_impact}

We now turn to studying how much engagement the politicians' posts generate and whether the use of metaphors may be associated with increased engagement. 
One of the core reasons politicians engage in social media is to engage with their constituents, and the larger electorate of the country. The level of engagement generated in terms of number of likes, comments etc. have been used in the past for measuring the level of engagement social media personalities manage to elicit from their followers \cite{guerini2012exploring,cvijikj2013online,muniz2019engagement,heiss2019drives}. For our study, we define five engagement metrics for each post:
\begin{itemize}
    \item \participation{}: audience engaging in discussion; measured $log(x+1)$ where $x$ is the number of direct comments in response to the post. 
    \item \propagation{}: audience propagating the message in the post; measured $log(x+1)$ where $x$ is the no. of times the post was shared.
    \item \acceptance{}: audience accepting or agreeing with the post, measured as $log(x+1)$ where $x$ is the no. of \textit{Like} reactions the post received.
    \item \posprovocation{}: $log(x+1)$ where $x$ is the number of positive reactions (\textit{Love}, \textit{Haha}, \textit{Wow}).
    \item \negprovocation{}: $log(x+1)$ where $x$ is the number of negative reactions (\textit{Angry}, \textit{Sad}).
\end{itemize}

\noindent Note that we use the $log(x+1)$ value for each engagement metric in order to account for zero-metaphoricity. Since Facebook launched the \textit{Love}, \textit{Haha}, \textit{Wow}, \textit{Angry}, and \textit{Sad} reactions only in Feb 2016, we limit any analysis involving \posprovocation{} and \negprovocation{} to only those posts authored on or after 01 Mar '16.
\noindent 

\subsection{Post-level Engagement Analysis}

\begin{table}
\centering
\begin{tabular}{lc}
\toprule
          Metric & Coefficient of Metaphoricity \\ 
\midrule
    \participation{} & 0.19\textbf{***} \\ 
    \propagation{} & 0.12\textbf{***}  \\ 
    \acceptance{} &  0.13\textbf{***} \\ 
    \posprovocation{} &  0.08\textbf{*}  \\ 
    \negprovocation{} &  0.10\textbf{*} \\ 
\bottomrule
\end{tabular}
\caption{\label{tab_impact_post_level} Coefficients obtained on mixed-effects model analysis at post level, controlling for fixed effects for post length, gender, party, and random effect for each politician.
Statistical Significance values (*: $p<0.05$; **: $p<0.01$; ***: $p<0.001$) reported after Bonferroni correction.}

\end{table}

Do posts with higher metaphoricity 
result in more engagement from readers? 
Figure~\ref{fig:fig_lmplots} shows the regression plots of post-level engagement features with respect to metaphoricity of posts. To control for the post length (since longer substantive posts may generate more engagement), we normalized the metaphoricity value by dividing them by the number of words in the posts. As can be seen from the figure, metaphoricity has a significant positive association with all three engagement features: \participation{}, \propagation{}, and \acceptance{}. However, the \posprovocation{} and \negprovocation{} features did not show a significant association, which may in part be due to the limited statistical power for those analyses, as described above.

However, there are other factors such as gender and party of the politician that may affect this relationship. Furthermore, factors associated with individual politicians (e.g., follower network size, approval rating etc.) may also impact the engagement their posts generate. In order to account for this correctly, we did two things: 1) balancing our data for this analysis by choosing a random sample of 100 posts from each politician who has made that many posts, and 2) accounting for group-level effects in each politician's posts by employing a Linear Mixed Effects \cite{lindstrom1988newton} analysis that allows random intercepts for each group. Linear mixed models allow both fixed and random effects, that accounts for the non independence in our data: multiple posts are made by the same politician. Table~\ref{tab_impact_post_level} shows the coefficients obtained on the mixed-effects model analysis at post level, controlling for fixed effects for post length, gender, party, and adding a random effect for each politician. Significance values (*: $p<0.05$; **: $p<0.01$; ***: $p<0.001$) are reported after Bonferroni correction.

\begin{figure*}[t]
    \centering
    \includegraphics[width=.7\linewidth]{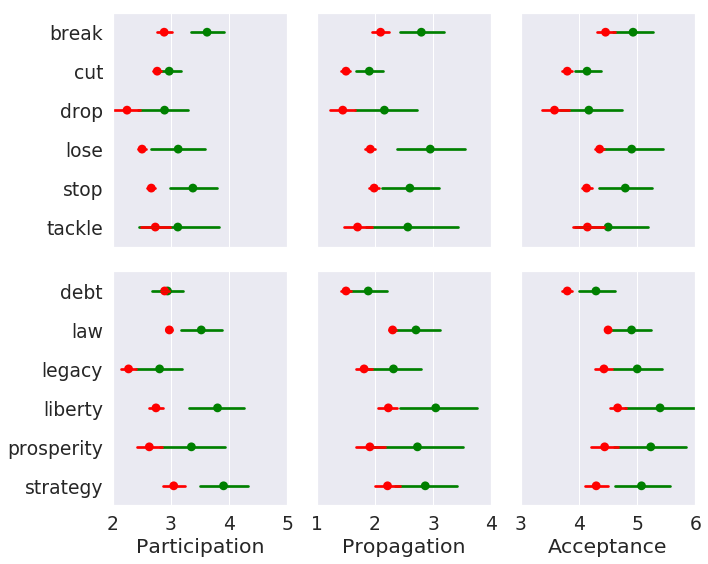}
    \caption{\small Point plots of Engagement features for some verbs and nouns chosen from those with the highest difference across metaphorical (green) vs. literal (red) use.}
    \label{fig:lemmaa_plot}
\end{figure*}



We find that despite the controls, metaphoricity of posts has significant positive associations with all engagement metrics. The largest coefficient is for \participation{} followed by \acceptance{} and \propagation{}. Note that since the engagement metrics are in log scale, a coefficient of .19 for \participation{} means: per unit increase in metaphoricity, we expect around 21\% ($ e^{0.19} = 1.21 $) increase in the number of comments the post receives.
Despite the fewer data points
(and hence less statistical power), \posprovocation{} and \negprovocation{} also showed smaller, but significant positive association with metaphoricity. 

\subsection{Word-level Engagement Analysis}
\label{sec_word_level}

While the previous section finds that posts with metaphors are associated with higher engagement, it does not prove that the increased engagement is owing to the metaphorical use alone. 
For instance, it may be the case that some words have high engagement potential (due to the topics they are associated with, e.g., \textit{law}, \textit{economy}) and they happen to be often used in metaphorical contexts.
To account for this issue, we need to compare the engagements in posts where a word is used metaphorically vs. where it is used literally. 

To do this, we first identify all lemmas in our corpus that were used in at least 10 posts metaphorically and 10 posts literally. This ensures a sufficient number of occurrences ($N>20$) per lemma, that enables the multi-level analysis we wish to perform.  
We exclude posts with multiple metaphors in order to eliminate effects from other metaphors. This results in a set of 513 lemmas and their instances in over 70K posts. 
Each such lemma instance, along with information about the post (length, engagement) and the politician who made the post (gender, party, etc.) forms the dataset for this analysis.

\begin{table}
\centering
\begin{tabular}{llcl}
\toprule
          Metric &               Verbs & Adjectives &               Nouns \\
\midrule
   \participation &  0.20\textbf{***} &       0.08 &  0.10\textbf{***} \\
     \propagation &  0.12\textbf{***} &      -0.01 &                0.05\textbf{*} \\
       \acceptance &  0.12\textbf{***} &      -0.02 &  0.10\textbf{***} \\
\posprovocation &                0.04 &      -0.05 &                0.07 \\
 \negprovocation &                0.03 &      -0.11 &                0.07 \\
\bottomrule

\end{tabular}
\caption{ \label{tab_impact_lemma_level} Coefficients on mixed effects linear model at lemma level, controlling for post length, gender, party, and random effect for each lemma (similar results with random effect for each politician).
Significance values (*: $p<0.05$; **: $p<0.01$; ***: $p<0.001$) after Bonferroni correction.}
\end{table}

Each lemma could take part in the metaphors as either a verb, an adjective, or a noun. We separated out the lemmas according to the syntactic role they play. When a word takes part in a metaphor as a verb or an adjective, it is evoking the framing in the source domain, whereas a noun taking part in a metaphor is usually representing the target domain. 
In other words, verbs and adjectives are the source domain words of the metaphor (i.e. they are themselves \textit{used} metaphorically) and nouns are the target domain words (i.e., they are \textit{described} metaphorically by the verbs or adjectives).
We first analyze whether our findings at the post level hold at the lemma level. 
For verbs and adjectives, this tests whether their metaphorical sense generates more engagement than the literal sense of the same word. For nouns, this tests whether they evoke more engagement when described metaphorically, than when described literally. 

As Table~\ref{tab_impact_lemma_level} shows, we find that when target domains (i.e., nouns) are described metaphorically, they generate significantly more engagement in \participation{}, \propagation{}, and \acceptance{}. In the source domain, only verbs, not adjectives, when used in their metaphorical sense result in increased engagement. We do not see any effect on \posprovocation{} or \negprovocation{}; it could partly be due to the smaller dataset size (less than 50 lemmas qualified for these two analyses).



Next, we inspect how individual words' engagement scores differ depending on whether they are used in metaphorical vs. literal sense. 
Figure~\ref{fig:lemmaa_plot} shows the point plots for \participation{}, \propagation{}, and \acceptance{} for a few verbs and nouns with the most difference (chosen from the top-10 in each category). We don't show examples of adjectives, as well as \posprovocation{} and \negprovocation{}, since we didn't find any overall effects in them.

The plots in the top row show source domain words (verbs) that have higher engagement in their metaphorical sense (green), than in their literal sense (red).
One common theme we observe in these verbs is that they all come from source domains related to physical experiences, associated with motion and movement (\textit{stop, drop}), mechanisms (\textit{break}) and wholeness (\textit{cut}).
The bottom row shows the examples for target domain words (nouns).
In contrast to verbs, nouns with the most difference in engagement metrics exemplify abstract target domains, relating to key topics of political discourse, such as values (\textit{liberty}), political process (\textit{strategy, law}) and finances (\textit{prosperity, debt}). When described metaphorically, as opposed to literally, these target domain words exhibit a significantly greater engagement, further supporting our findings on the impact of metaphor in political discourse. 

\section{Discussion}
\label{sec_discussion}

In this paper, we demonstrated the positive effect of metaphor use in generating engagement in political discourse. We used an NLP-based metaphor detection approach that allowed us to perform a large-scale data-driven topic-agnostic study, unlike previous work on this topic. Our results hold true despite controlling for gender and party affiliation of politicians. Furthermore, since longer posts are more likely to contain more interesting content that may drive engagement, we control for post length (i.e., the number of words) in all our analyses. This way, we ensure that our results hold true regardless of the length of the posts. Where applicable, we also control for the random effects due to individual politicians using a mixed effects model to ensure that the increased engagement we find in posts with more metaphoricity is not driven by a handful of politicians who use more metaphors or happen to have more engaging followers. These controls are also applied to our word-level engagement analysis, which goes one step further to test the difference in engagement when the same word is used literally vs. metaphorically. 

In addition to the above controls, we performed further post-hoc analyses to eliminate other potential confounds. First, we already found that metaphor use was higher after the elections (especially for Democrats), and it is possible that the election loss could have caused high engagement. So the association we find between engagement and metaphoricity may be mediated by the election result (i.e., both positively impacted by the election). To ensure that this is not solely driving our findings, we first re-ran our analyses only on posts authored prior to election results, and verified that our findings do indeed hold true for that subset of the data. To further understand the effect of election results, we introduced a binary variable denoting whether a post was made prior to or after the election as an extra control into our post-level mixed effects analyses. As expected, we found that the engagement metrics had a positive association with whether the post was made after the elections (e.g., $b=1.225$, $t(0.066)$, $p<0.001$ for \participation{}). However, our findings on the impact of metaphoricity on engagement metrics (Table 1) stayed stable across all analyses (i.e., direction of relationship and statistical significance), although the effect size was reduced marginally. We did not find any interaction effects between these variables. In other words, while the posts made after the election resulted in higher engagement, the relationship between metaphoricity and engagement metrics holds true regardless of whether the post was made before or after elections. 

Another interesting question is whether the engagement patterns in response to metaphoricity differs based on the posters' political affiliation. In order to test this, we tested for interaction effects between political affiliation of the poster and metaphoricity of the post on the engagement metrics (retaining the other controls as in previous analyses). We found no significant interaction effects between the two. In other words, political affiliation do not have a significant effect on how metaphoricity impacts engagement. 



\section{Conclusion}
\label{sec_conclusion}

We presented a large-scale domain-agnostic study of the use and impact of metaphors in political discourse, using a dataset of Facebook posts made by US politicians. We showed that posts with metaphors exhibit higher engagement from their audience. We further performed a finer-grained analysis of over 500 words' metaphorical vs. literal usage across 70K posts to demonstrate that metaphorical uses/expressions do indeed generate more engagement than the corresponding literal ones. We also found interesting nuances in how words of different parts of speech tags, and of different domains/topics pattern differently in their engagement potential with regard to metaphoricity. 
To our knowledge, this is the first large-scale study of the efficacy of metaphor use in eliciting engagement in political discourse. 

Our topic-agnostic method of studying metaphor use in political discourse may aid in further studies on how politicians' speech can evoke actions from their followers. While we focused on the online engagement metrics measured through shares, comments and reactions, future work could also study how metaphors may evoke offline engagements and actions in followers, such as turning out for elections, violent protests and insurrections etc. This may be an especially important aspect to study, as politicians across the world are increasingly using social media to interact with and influence their electorate. 

Future work may also look further into the semantic and topical regularities in these phenomena, as we observed in Figure~\ref{fig:lemmaa_plot}. Understanding what kinds of metaphors are likely to elicit higher engagement might be of importance to political scientists and strategists. Such word level analyses will also aid in studying how certain types of metaphors might evoke certain kinds of offline reactions in followers; for instance, are certain kinds of metaphors more likely to evoke the followers to turn to violent acts offline? 

Another direction of future work is to extend the study to other domains, such as brand management and advertisement, to see if the patterns we observe in this study do transfer to other domains. Future work should also look into whether these findings extend to other geographies, cultures, and languages. 






\bibliography{paper,anthology}

\begin{thebibliography}{50}
\providecommand{\natexlab}[1]{#1}
\providecommand{\url}[1]{\texttt{#1}}
\providecommand{\urlprefix}{URL }
\expandafter\ifx\csname urlstyle\endcsname\relax
  \providecommand{\doi}[1]{doi:\discretionary{}{}{}#1}\else
  \providecommand{\doi}{doi:\discretionary{}{}{}\begingroup
  \urlstyle{rm}\Url}\fi

\bibitem[{Beigman~Klebanov et~al.(2016)Beigman~Klebanov, Leong, Gutierrez,
  Shutova, and Flor}]{BeigmanKlebanovACL2016}
Beigman~Klebanov, B.; Leong, C.~W.; Gutierrez, E.~D.; Shutova, E.; and Flor, M.
  2016.
\newblock Semantic classifications for detection of verb metaphors.
\newblock In \emph{Proceedings of ACL}, 101--106. Berlin, Germany.

\bibitem[{Bulat, Clark, and Shutova(2017)}]{Bulat2017}
Bulat, L.; Clark, S.; and Shutova, E. 2017.
\newblock {Modelling metaphor with attribute-based semantics}.
\newblock \emph{EACL 2017} .

\bibitem[{Cameron(2003)}]{cameron2003metaphor}
Cameron, L. 2003.
\newblock \emph{Metaphor in educational discourse}.
\newblock A\&C Black.

\bibitem[{Citron and Goldberg(2014)}]{citron2014metaphorical}
Citron, F.~M.; and Goldberg, A.~E. 2014.
\newblock Metaphorical sentences are more emotionally engaging than their
  literal counterparts.
\newblock \emph{Journal of cognitive neuroscience} 26(11): 2585--2595.

\bibitem[{Crawford(2009)}]{Crawford2009}
Crawford, E. 2009.
\newblock {Conceptual Metaphors of Affect}.
\newblock \emph{Emotion Review} 1(2).

\bibitem[{Cvijikj and Michahelles(2013)}]{cvijikj2013online}
Cvijikj, I.~P.; and Michahelles, F. 2013.
\newblock Online engagement factors on Facebook brand pages.
\newblock \emph{Social network analysis and mining} 3(4): 843--861.

\bibitem[{Danescu-Niculescu-Mizil et~al.(2012)Danescu-Niculescu-Mizil, Cheng,
  Kleinberg, and Lee}]{danescu2012you}
Danescu-Niculescu-Mizil, C.; Cheng, J.; Kleinberg, J.; and Lee, L. 2012.
\newblock You had me at hello: How phrasing affects memorability.
\newblock In \emph{Proceedings of the ACL}.

\bibitem[{Danescu-Niculescu-Mizil et~al.(2009)Danescu-Niculescu-Mizil,
  Kossinets, Kleinberg, and Lee}]{danescu2009opinions}
Danescu-Niculescu-Mizil, C.; Kossinets, G.; Kleinberg, J.; and Lee, L. 2009.
\newblock How opinions are received by online communities: a case study on
  amazon. com helpfulness votes.
\newblock In \emph{Proceedings of the 18th international conference on World
  wide web}, 141--150.

\bibitem[{Dankers et~al.(2019)Dankers, Rei, Lewis, and Shutova}]{Dankers2019}
Dankers, V.; Rei, M.; Lewis, M.; and Shutova, E. 2019.
\newblock Modelling the interplay of metaphor and emotion through multitask
  learning.
\newblock In \emph{Proceedings of EMNLP 2019}. Hong Kong.

\bibitem[{Dunn(2013)}]{DunnCicling2013}
Dunn, J. 2013.
\newblock Evaluating the premises and results of four metaphor identification
  systems.
\newblock In \emph{Proceedings of CICLing'13}, 471--486. Samos, Greece.

\bibitem[{Entman(2003)}]{Entman2003}
Entman, R. 2003.
\newblock {Language: The Loaded Weapon}.
\newblock \emph{Political Communication} (20): 415--432.

\bibitem[{Gao et~al.(2018)Gao, Choi, Choi, and Zettlemoyer}]{gao2018neural}
Gao, G.; Choi, E.; Choi, Y.; and Zettlemoyer, L. 2018.
\newblock Neural {M}etaphor {D}etection in {C}ontext.
\newblock In \emph{Conference on {E}mpirical {M}ethods in {N}atural {L}anguage
  {P}rocessing}.
\newblock \doi{10.18653/v1/D18-1060}.

\bibitem[{Guerini, Strapparava, and Ozbal(2012)}]{guerini2012exploring}
Guerini, M.; Strapparava, C.; and Ozbal, G. 2012.
\newblock Exploring text virality in social networks.
\newblock \emph{arXiv preprint arXiv:1203.5502} .

\bibitem[{Guerini, Strapparava, and Stock(2008)}]{guerini2008trusting}
Guerini, M.; Strapparava, C.; and Stock, O. 2008.
\newblock Trusting politicians’ words (for persuasive NLP).
\newblock In \emph{International Conference on Intelligent Text Processing and
  Computational Linguistics}, 263--274. Springer.

\bibitem[{Guti{\'{e}}rrez et~al.(2016)Guti{\'{e}}rrez, Shutova, Marghetis, and
  Bergen}]{DarioACL2016}
Guti{\'{e}}rrez, E.~D.; Shutova, E.; Marghetis, T.; and Bergen, B. 2016.
\newblock Literal and Metaphorical Senses in Compositional Distributional
  Semantic Models.
\newblock In \emph{Proceedings of the 54th Annual Meeting of the Association
  for Computational Linguistics, {ACL} 2016, August 7-12, 2016, Berlin,
  Germany, Volume 1: Long Papers}.

\bibitem[{Heintz et~al.(2013)Heintz, Gabbard, Srivastava, Barner, Black,
  Friedman, and Weischedel}]{heintz-EtAl:2013:Meta4NLP}
Heintz, I.; Gabbard, R.; Srivastava, M.; Barner, D.; Black, D.; Friedman, M.;
  and Weischedel, R. 2013.
\newblock Automatic Extraction of Linguistic Metaphors with LDA Topic Modeling.
\newblock In \emph{Proceedings of the First Workshop on Metaphor in NLP},
  58--66. Atlanta, Georgia.
\newblock \urlprefix\url{http://www.aclweb.org/anthology/W13-0908}.

\bibitem[{Heiss, Schmuck, and Matthes(2019)}]{heiss2019drives}
Heiss, R.; Schmuck, D.; and Matthes, J. 2019.
\newblock What drives interaction in political actors’ Facebook posts?
  Profile and content predictors of user engagement and political actors’
  reactions.
\newblock \emph{Information, Communication \& Society} 22(10): 1497--1513.

\bibitem[{Honnibal and Johnson(2015)}]{honnibal-johnson:2015:EMNLP}
Honnibal, M.; and Johnson, M. 2015.
\newblock An Improved Non-monotonic Transition System for Dependency Parsing.
\newblock In \emph{Proceedings of the 2015 Conference on Empirical Methods in
  Natural Language Processing}, 1373--1378. Lisbon, Portugal: Association for
  Computational Linguistics.
\newblock \urlprefix\url{https://aclweb.org/anthology/D/D15/D15-1162}.

\bibitem[{Jang(2017)}]{jang2017computational}
Jang, H. 2017.
\newblock \emph{Computational Modeling of Metaphor in Discourse}.
\newblock Ph.D. thesis, Carnegie Mellon University.

\bibitem[{Jang et~al.(2017)Jang, Maki, Hovy, and Rose}]{jang2017finding}
Jang, H.; Maki, K.; Hovy, E.; and Rose, C. 2017.
\newblock Finding structure in figurative language: Metaphor detection with
  topic-based frames.
\newblock In \emph{Proceedings of the 18th Annual SIGdial Meeting on Discourse
  and Dialogue}, 320--330.

\bibitem[{Lakkaraju, McAuley, and Leskovec(2013)}]{lakkaraju2013s}
Lakkaraju, H.; McAuley, J.~J.; and Leskovec, J. 2013.
\newblock What's in a Name? Understanding the Interplay between Titles,
  Content, and Communities in Social Media.
\newblock \emph{ICWSM} 1(2): 3.

\bibitem[{Lakoff(1991)}]{Lakoff1991}
Lakoff, G. 1991.
\newblock {Metaphor and war: The metaphor system used to justify war in the
  Gulf}.
\newblock \emph{Peace Research} 23: 25--32.

\bibitem[{Lakoff(2002)}]{LakoffMoralPolitics}
Lakoff, G. 2002.
\newblock \emph{Moral Politics: How Liberals and Conservatives Think}.
\newblock University of Chicago Press.

\bibitem[{Lakoff and Johnson(1980)}]{LakoffAndJohnson}
Lakoff, G.; and Johnson, M. 1980.
\newblock \emph{Metaphors We Live By}.
\newblock Chicago: University of Chicago Press.

\bibitem[{Lakoff and Wehling(2012)}]{LakoffWehling}
Lakoff, G.; and Wehling, E. 2012.
\newblock \emph{The Little Blue Book: The Essential Guide to Thinking and
  Talking Democratic}.
\newblock New York: Free Press.

\bibitem[{Lindstrom and Bates(1988)}]{lindstrom1988newton}
Lindstrom, M.~J.; and Bates, D.~M. 1988.
\newblock Newton—Raphson and EM algorithms for linear mixed-effects models
  for repeated-measures data.
\newblock \emph{Journal of the American Statistical Association} 83(404):
  1014--1022.

\bibitem[{Mikolov et~al.(2013)Mikolov, Chen, Corrado, and Dean}]{Mikolov2013a}
Mikolov, T.; Chen, K.; Corrado, G.; and Dean, J. 2013.
\newblock {Efficient Estimation of Word Representations in Vector Space}.
\newblock In \emph{Proceedings of the International Conference on Learning
  Representations (ICLR 2013)}.
\newblock ISBN 9781577357384.
\newblock ISSN 10450823.
\newblock \doi{10.1162/153244303322533223}.
\newblock \urlprefix\url{http://arxiv.org/pdf/1301.3781v3.pdf}.

\bibitem[{Mohammad, Shutova, and Turney(2016)}]{mohammad2016metaphor}
Mohammad, S.; Shutova, E.; and Turney, P. 2016.
\newblock Metaphor as a medium for emotion: An empirical study.
\newblock In \emph{Proceedings of the Fifth Joint Conference on Lexical and
  Computational Semantics}, 23--33.

\bibitem[{Mohler et~al.(2013)Mohler, Bracewell, Tomlinson, and
  Hinote}]{mohler-EtAl:2013:Meta4NLP}
Mohler, M.; Bracewell, D.; Tomlinson, M.; and Hinote, D. 2013.
\newblock Semantic Signatures for Example-Based Linguistic Metaphor Detection.
\newblock In \emph{Proceedings of the First Workshop on Metaphor in NLP},
  27--35. Atlanta, Georgia.
\newblock \urlprefix\url{http://www.aclweb.org/anthology/W13-0904}.

\bibitem[{Mu{\~n}iz et~al.(2019)Mu{\~n}iz, Campos-Dom{\'\i}nguez, Saldierna,
  and Dader}]{muniz2019engagement}
Mu{\~n}iz, C.; Campos-Dom{\'\i}nguez, E.; Saldierna, A.~R.; and Dader, J.~L.
  2019.
\newblock Engagement of politicians and citizens in the cyber campaign on
  Facebook: a comparative analysis between Mexico and Spain.
\newblock \emph{Contemporary Social Science} 14(1): 102--113.

\bibitem[{Musolff(2000)}]{Musolff}
Musolff, A. 2000.
\newblock \emph{Mirror images of Europe: Metaphors in the public debate about
  Europe in Britain and Germany}.
\newblock Muenchen: Iudicium.

\bibitem[{Prabhakaran, Arora, and Rambow(2014)}]{prabhakaran-etal-2014-staying}
Prabhakaran, V.; Arora, A.; and Rambow, O. 2014.
\newblock Staying on Topic: An Indicator of Power in Political Debates.
\newblock In \emph{Proceedings of the 2014 Conference on Empirical Methods in
  Natural Language Processing ({EMNLP})}, 1481--1486. Doha, Qatar: Association
  for Computational Linguistics.
\newblock \doi{10.3115/v1/D14-1157}.
\newblock \urlprefix\url{https://www.aclweb.org/anthology/D14-1157}.

\bibitem[{Prabhakaran, John, and Seligmann(2013)}]{prabhakaran-etal-2013-upper}
Prabhakaran, V.; John, A.; and Seligmann, D.~D. 2013.
\newblock Who Had the Upper Hand? Ranking Participants of Interactions Based on
  Their Relative Power.
\newblock In \emph{Proceedings of the Sixth International Joint Conference on
  Natural Language Processing}, 365--373. Nagoya, Japan: Asian Federation of
  Natural Language Processing.
\newblock \urlprefix\url{https://www.aclweb.org/anthology/I13-1042}.

\bibitem[{Rei et~al.(2017)Rei, Bulat, Kiela, and Shutova}]{MarekEMNLP2017}
Rei, M.; Bulat, L.; Kiela, D.; and Shutova, E. 2017.
\newblock {Grasping the Finer Point: A Supervised Similarity Network for
  Metaphor Detection}.
\newblock In \emph{Proceedings of EMNLP 2017}.

\bibitem[{Rosenberg and Hirschberg(2009)}]{rosenberg2009charisma}
Rosenberg, A.; and Hirschberg, J. 2009.
\newblock Charisma perception from text and speech.
\newblock \emph{Speech Communication} 51(7): 640--655.

\bibitem[{Shutova(2010)}]{shutova2010models}
Shutova, E. 2010.
\newblock Models of metaphor in NLP.
\newblock In \emph{Proceedings of the 48th annual meeting of the association
  for computational linguistics}, 688--697. Association for Computational
  Linguistics.

\bibitem[{Shutova(2011)}]{ShutovaThesis}
Shutova, E. 2011.
\newblock \emph{Computational Approaches to Figurative Language}.
\newblock Ph.D. thesis, University of Cambridge, UK.

\bibitem[{Shutova(2013)}]{shutova-2013-metaphor}
Shutova, E. 2013.
\newblock Metaphor Identification as Interpretation.
\newblock In \emph{Second Joint Conference on Lexical and Computational
  Semantics (*{SEM}), Volume 1: Proceedings of the Main Conference and the
  Shared Task: Semantic Textual Similarity}, 276--285. Atlanta, Georgia, USA:
  Association for Computational Linguistics.
\newblock \urlprefix\url{https://www.aclweb.org/anthology/S13-1040}.

\bibitem[{Shutova, Kiela, and Maillard(2016)}]{ShutovaNAACL2016}
Shutova, E.; Kiela, D.; and Maillard, J. 2016.
\newblock Black Holes and White Rabbits: Metaphor Identification with Visual
  Features.
\newblock In \emph{Proceedings of NAACL-HLT 2016}. San Diego, CA.

\bibitem[{Steen et~al.(2010)Steen, Dorst, Herrmann, Kaal, Krennmayr, and
  Pasma}]{SteenMIPVU}
Steen, G.~J.; Dorst, A.~G.; Herrmann, J.~B.; Kaal, A.~A.; Krennmayr, T.; and
  Pasma, T. 2010.
\newblock \emph{{A method for linguistic metaphor identification: From MIP to
  MIPVU}}.
\newblock Amsterdam/Philadelphia: John Benjamins.

\bibitem[{Strzalkowski et~al.(2013)Strzalkowski, Broadwell, Taylor, Feldman,
  Shaikh, Liu, Yamrom, Cho, Boz, Cases, and
  Elliot}]{strzalkowski-EtAl:2013:Meta4NLP}
Strzalkowski, T.; Broadwell, G.~A.; Taylor, S.; Feldman, L.; Shaikh, S.; Liu,
  T.; Yamrom, B.; Cho, K.; Boz, U.; Cases, I.; and Elliot, K. 2013.
\newblock Robust Extraction of Metaphor from Novel Data.
\newblock In \emph{Proceedings of the First Workshop on Metaphor in NLP},
  67--76. Atlanta, Georgia.
\newblock \urlprefix\url{http://www.aclweb.org/anthology/W13-0909}.

\bibitem[{Tan, Lee, and Pang(2014)}]{tan2014effect}
Tan, C.; Lee, L.; and Pang, B. 2014.
\newblock The effect of wording on message propagation: Topic- and
  author-controlled natural experiments on {T}witter.
\newblock In \emph{Proceedings of the 52nd Annual Meeting of the Association
  for Computational Linguistics (Volume 1: Long Papers)}, 175--185. Baltimore,
  Maryland: Association for Computational Linguistics.
\newblock \doi{10.3115/v1/P14-1017}.
\newblock \urlprefix\url{https://www.aclweb.org/anthology/P14-1017}.

\bibitem[{Tannen(1993)}]{Tannen}
Tannen, D. 1993.
\newblock \emph{Framing in Discourse}.
\newblock Oxford, UK: Oxford University Press.

\bibitem[{Thibodeau and Boroditsky(2011)}]{ThibodeauBoroditsky}
Thibodeau, P.~H.; and Boroditsky, L. 2011.
\newblock Metaphors We Think With: The Role of Metaphor in Reasoning.
\newblock \emph{PLoS ONE} 6(2): e16782.

\bibitem[{Tsvetkov et~al.(2014)Tsvetkov, Boytsov, Gershman, Nyberg, and
  Dyer}]{Boytsov2014}
Tsvetkov, Y.; Boytsov, L.; Gershman, A.; Nyberg, E.; and Dyer, C. 2014.
\newblock {Metaphor Detection with Cross-Lingual Model Transfer}.
\newblock \emph{Proceedings of the 52nd Annual Meeting of the Association for
  Computational Linguistics (ACL 2014)} 248--258.
\newblock \doi{10.3115/v1/P14-1024}.

\bibitem[{Turney et~al.(2011)Turney, Neuman, Assaf, and Cohen}]{Turney2011}
Turney, P.~D.; Neuman, Y.; Assaf, D.; and Cohen, Y. 2011.
\newblock Literal and metaphorical sense identification through concrete and
  abstract context.
\newblock In \emph{Proceedings of the Conference on Empirical Methods in
  Natural Language Processing}, EMNLP '11, 680--690. Stroudsburg, PA, USA:
  Association for Computational Linguistics.

\bibitem[{Voigt et~al.(2018)Voigt, Jurgens, Prabhakaran, Jurafsky, and
  Tsvetkov}]{voigt-etal-2018-rtgender}
Voigt, R.; Jurgens, D.; Prabhakaran, V.; Jurafsky, D.; and Tsvetkov, Y. 2018.
\newblock {R}t{G}ender: A Corpus for Studying Differential Responses to Gender.
\newblock In \emph{Proceedings of the Eleventh International Conference on
  Language Resources and Evaluation ({LREC}-2018)}. Miyazaki, Japan: European
  Languages Resources Association (ELRA).
\newblock \urlprefix\url{https://www.aclweb.org/anthology/L18-1445}.

\bibitem[{Wilks et~al.(2013)Wilks, Dalton, Allen, and
  Galescu}]{wilks-EtAl:2013:Meta4NLP}
Wilks, Y.; Dalton, A.; Allen, J.; and Galescu, L. 2013.
\newblock Automatic Metaphor Detection using Large-Scale Lexical Resources and
  Conventional Metaphor Extraction.
\newblock In \emph{Proceedings of the First Workshop on Metaphor in NLP},
  36--44. Atlanta, Georgia.
\newblock \urlprefix\url{http://www.aclweb.org/anthology/W13-0905}.

\bibitem[{Yang et~al.(2019)Yang, Chen, Yang, Jurafsky, and Hovy}]{yang2019let}
Yang, D.; Chen, J.; Yang, Z.; Jurafsky, D.; and Hovy, E. 2019.
\newblock Let’s make your request more persuasive: Modeling persuasive
  strategies via semi-supervised neural nets on crowdfunding platforms.
\newblock In \emph{Proceedings of the 2019 Conference of the North American
  Chapter of the Association for Computational Linguistics: Human Language
  Technologies, Volume 1 (Long and Short Papers)}, 3620--3630.

\bibitem[{Zeiler(2012)}]{Zeiler2012}
Zeiler, M.~D. 2012.
\newblock {ADADELTA: An Adaptive Learning Rate Method}.
\newblock \emph{arXiv preprint arXiv:1212.5701}
  \urlprefix\url{http://arxiv.org/abs/1212.5701}.

\end{thebibliography}

\end{document}